# A Function Approximation Approach to Estimation of Policy Gradient for POMDP with Structured Policies

Huizhen Yu
Lab for Information and Decision Systems
Massachusetts Institute of Technology
Cambridge, MA 02139

## Abstract

We consider the estimation of the policy gradient in partially observable Markov decision processes (POMDP) with a special class of structured policies that are finite-state controllers. We show that the gradient estimation can be done in the Actor-Critic framework, by making the critic compute a "value" function that does not depend on the states of POMDP. This function is the conditional mean of the true value function that depends on the states. We show that the critic can be implemented using temporal difference (TD) methods with linear function approximations, and the analytical results on TD and Actor-Critic can be transferred to this case. Although Actor-Critic algorithms have been used extensively in Markov decision processes (MDP), up to now they have not been proposed for POMDP as an alternative to the earlier proposal GPOMDP algorithm, an actor-only method. Furthermore, we show that the same idea applies to semi-Markov problems with a subset of finite-state controllers.

## 1 INTRODUCTION

We consider discrete-time infinite-horizon partially observable Markov decision processes (POMDP) with finite spaces of states, controls and observations. The model is as follows. Let $X_t$ be the state, $Y_t$ the observation, and $U_t$ the control at time $t$. The per-stage cost is $g_t = g(X_t, Y_t, U_t)$. At time $t+1$, given $\{(X_k, Y_k, U_k)\}_{k \leq t}$, the system transits to the state $X_{t+1}$ with probability $p(X_{t+1}|X_t, U_t)$, and the state then generates the observation $Y_{t+1}$ with probability $p(Y_{t+1}|X_{t+1}, U_t)$. The states $X_t$ are not observable. Controls are chosen with knowledge of the past observations, controls, and/or the distribution of the initial state.

We limit the policy space to the set of finite-state controllers. A finite-state controller is like a probabilistic automaton, with the observations being its inputs and the controls its outputs. The controller has a finite number of "internal-states" that evolve in a Markovian way, and it outputs a control depending on the current internal state and the current observation. We consider the average cost criterion.

The finite-state controller approach to POMDP has been proposed in the work of "GPOMDP" (Baxter and Bartlett, 2001), and "Internal-state POMDP" (Aberdeen and Baxter, 2002). There are two distinctive features about finite-state controllers. One is that the state of POMDP, (even though not observable), the observation, and the internal state of the controller jointly form a Markov process, so the theory of finite-state Markov decision processes (MDP) applies. In contrast, the asymptotic behavior of POMDP under a general policy is much harder to establish. The other distinctive feature of finite-state controllers is that the gradient of the cost with respect to the policy parameters can be estimated from sample trajectories, without requiring the explicit model of POMDP, so gradient-based methods can be used for policy improvement. This feature is appealing for both large problems in which either models are not represented explicitly, or exact inferences are intractable, and reinforcement learning problems in which the environment model is unknown and may be varying in time.

As the states of POMDP are not observable, the gradient estimation method proposed by (Baxter and Bartlett, 2001) and (Aberdeen and Baxter, 2002) avoids estimating the value function. The idea there is to replace the value of a state in the gradient expression by the path-dependent random cost starting from that state. To our knowledge, up to now gradient estimators that use a value function approximator have not been proposed as an alternative to GPOMDP in learning finite-state controllers for POMDP.[1] To propose such an alternative is the purpose of this paper.

We show that the gradient is computable by a function approximation approach. Without pre-committing to a specific estimation algorithm, we start with rewriting the gradient expression so that it involves a "value" function that does not depend on the states. This "value" function is the conditional mean of the true value function given the observation, action, internal states under the equilibrium distribution of the Markov chain. By ergodicity, biased estimates of this "value" function can be obtained from sample trajectories. In particular, temporal difference (TD) methods with linear function approximation, including both $\beta$-discounted TD($\lambda$) and average cost TD($\lambda$), can be used, and the biases of the corresponding gradient estimators asymptoticly go to zero when $\beta \to 1, \lambda \to 1$.

The computation of this value function may be viewed as the critic part of the actor-critic framework (e.g., (Konda, 2002)), in which the critic evaluates the policy, and the actor improves the policy based on the evaluation. Thus for POMDP with finite-state controllers, the algorithms as well as their analysis fit in the general MDP methodology with both actor-only and actor-critic methods, and can be viewed as special cases.

The idea of estimating the conditional mean of the true value function first appeared in (Jaakkola, Singh, and Jordan, 1994), which is a gradient-descent flavored method in the context of the finite memory approach to reinforcement learning in POMDP. This earlier work does not start with gradient estimation, though it is closely related to. Our way of using this idea in rewriting the gradient expression is to some degree new. Algorithmically one does not have to take this additional conditional expectation step in order to apply the Actor-Critic framework. However, making this condition mean explicit in the gradient expression, we think, is a more direct approach, and allows the use of other estimation algorithms such as non-linear function approximators.

Finally we show that the same function approximation approach also applies to gradient estimation in semi-Markov problems, for which an earlier proposal is a GPOMDP type algorithm (Singh, Tadic, and Doucet, 2002).

---

[1](Meuleau et al., 1999) uses the value function to parameterize the policy and uses the path-dependent random cost for gradient estimation in episodic settings. A GPOMDP/SARSA hybrid was proposed by Aberdeen and Baxter in an early work. However, the reasoning there was incorrect, because the marginal process of internal-state and observation is not a Markov chain.

The paper is organized as follows. In Section 2, we lay out our approach for reactive policies, a simple subclass of finite-state controllers, yet captures all the main ideas in the analysis. We introduce the background, and present the gradient expressions, the algorithms, and an error analysis. In Section 3, we present the gradient estimation algorithm for finite-state controllers, and in Section 4, for semi-Markov problems. In Section 5, we provide experiments, showing that the estimates using function approximation are comparable to those from an improved GPOMDP method that uses a simple variance reduction technique, and in addition the function approximation approach provides more options in controlling bias and variance.

## 2 GRADIENT ESTIMATION FOR REACTIVE POLICIES

We first present our approach for the simplest finite-state controllers – reactive policies, mainly for their notational simplicity. We will also introduce the background of policy gradient estimation. A reactive policy is a randomized stationary policy such that the probability of taking a control is a function of the most recent observation only. The graphical model of POMDP with a reactive policy is shown in Fig. 1. The process $\{(X_t, Y_t, U_t)\}$ jointly forms a Markov chain under a reactive policy, and so does the *marginal* process $\{(X_t, Y_t)\}$, (marginalized over controls $U_t$).

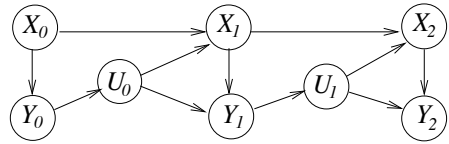

Figure 1: POMDP with a Reactive Policy.

Let $\{\gamma_\theta \mid \theta \in \Theta\}$ be a family of reactive policies parametrized by $\theta$. For any policy $\gamma_\theta$, let

$$\mu_u(y, \theta) = p(U_t = u \mid Y_t = y; \theta)$$

be the probability of taking control $u$ upon the observation $y$. The following assumptions are standard. We require that $\mu_u(y, \theta)$ is differentiable for any given $u, y$, and the transition probability $p(X_{t+1} = \bar{x}, Y_{t+1} = \bar{y} \mid X_t = x, Y_t = y; \theta)$ is differentiable for any given $x, y, \bar{x}, \bar{y}$. Furthermore we assume that for all $\theta \in \Theta$ the Markov chains $\{(X_t, Y_t)\}$ are defined on a common state space[2] and

**Assumption 1** *Under any policy $\gamma_\theta$, the Markov chain $\{(X_t, Y_t)\}$ is irreducible and aperiodic.*

---

[2]Note this is not necessarily the product of the state and the observation space of POMDP.

**Assumption 2** *There exists a constant $L$, such that for all $\theta \in \Theta$, $\max_{u,y} \left\| \frac{\nabla \mu_u(y,\theta)}{\mu_u(y,\theta)} \right\| \leq L$, where $0/0$ is regarded as $0$.*

The first assumption ensures that the average cost is a constant and differentiable for all policy $\gamma_\theta$. (A short proof of differentiability in finite-state MDPs in general is given in the appendix.[3]) The second assumption of boundedness is to make it possible to compute the gradient by sampling methods.

## 2.1 REVIEW OF GRADIENT AND GRADIENT APPROXIMATION

Let $\eta(\theta)$ be the average cost of the reactive policy $\gamma_\theta$, and let $E_0^\theta$ denote expectation[4] with respect to the *equilibrium distribution* of the Markov chain $\{(X_t, Y_t, U_t)\}$ under policy $\gamma_\theta$. For simplicity of notation, we will drop $\theta$ in $\eta(\theta)$ and $E_0^\theta$, and use $\eta$ and $E_0$ throughout the paper.

Suppose $\theta = (\theta_1, \ldots, \theta_k) \in \mathcal{R}^k$, and let $\nabla \mu_u(y, \theta)$ be the column vector $\left( \frac{\partial \mu_u(y,\theta)}{\partial \theta_1}, \ldots, \frac{\partial \mu_u(y,\theta)}{\partial \theta_k} \right)'$. The gradient $\nabla \eta$ can be expressed as

$$\nabla \eta = E_0 \left\{ \frac{\nabla \mu_U(Y,\theta)}{\mu_U(Y,\theta)} Q(X,Y,U) \right\}, \qquad (1)$$

where the Q-function $Q(x,y,u)$ is defined by

$$Q(x,y,u) = g(x,y,u) + E\{h(X_1, Y_1) \mid X_0 = x, Y_0 = y, U_0 = u\},$$

and $h(x,y)$ is the bias (Puterman, 1994), defined by

$$h(x,y) = \lim_{T \to \infty} E \left\{ \sum_{t=0}^{T} (g_t - \eta) \mid X_0 = x, Y_0 = y \right\}.$$

In order to compute $h(x,y)$ directly one needs to pick a particular pair $(x_0, y_0)$ as a regenerating state (Marbach and Tsitsiklis, 2001). Since this is not possible in POMDP, one has to approximate it by other terms.

(Baxter and Bartlett, 2001) proposed the following approximate gradient:

$$\nabla_\beta \eta \stackrel{def}{=} E_0 \left\{ \frac{\nabla \mu_U(Y,\theta)}{\mu_U(Y,\theta)} Q_\beta(X,Y,U) \right\}, \qquad (2)$$

where $Q_\beta$ is the Q-function of the discounted problem:

$$Q_\beta(x,y,u) = g(x,y,u) + \beta E\{J_\beta(X_1, Y_1) \mid X_0 = x, Y_0 = y, U_0 = u\},$$

---
[3]Appendices of this paper can be found at the author's website (http://www.mit.edu/~janey).

[4]Conditional expectations are defined in the same way, i.e., with respect to the conditional distributions from the equilibrium distribution.

---

and $J_\beta$ is the cost function of the $\beta$-discounted problem. The approximate gradient $\nabla_\beta \eta$ converges to $\nabla \eta$ when $\beta \uparrow 1$. This is due to the fact that when $\beta \approx 1$, $J_\beta - \frac{\eta}{1-\beta} \approx h$, and $E_0\{\frac{\nabla \mu_U(Y,\theta)}{\mu_U(Y,\theta)}\} = 0$, therefore $E_0\{\frac{\nabla \mu_U(Y,\theta)}{\mu_U(Y,\theta)} c\} = 0$ for any constant $c$. Although the state is not observable, an estimate of $Q_\beta(X_t, Y_t, U_t)$ can be obtained by accumulating the costs along the *future* sample path starting from time $t$. This is the idea of the GPOMDP algorithm that estimates the approximate gradient $\nabla_\beta \eta$ by a sampling version of Eq. (2).

## 2.2 A NEW GRADIENT EXPRESSION FOR ESTIMATION

We first write Eq. (1) in a different way:

$$\begin{aligned} \nabla \eta &= E_0 \left\{ \frac{\nabla \mu_U(Y,\theta)}{\mu_U(Y,\theta)} Q(X,Y,U) \right\} \\ &= E_0 \left\{ \frac{\nabla \mu_U(Y,\theta)}{\mu_U(Y,\theta)} E_0\{Q(X,Y,U) \mid Y, U\} \right\} \\ &= E_0 \left\{ \frac{\nabla \mu_U(Y,\theta)}{\mu_U(Y,\theta)} v(Y,U) \right\}, \end{aligned} \qquad (3)$$

where

$$v(Y,U) = E_0 \{Q(X,Y,U) \mid Y, U\}, \qquad (4)$$

a function that depends on observation and action only. Similarly define $v_\beta(Y,U)$ to be the conditional mean of $Q_\beta$ given $Y, U$, and the approximate gradient (Eq. (2)) can be written as

$$\nabla_\beta \eta = E_0 \left\{ \frac{\nabla \mu_U(Y,\theta)}{\mu_U(Y,\theta)} v_\beta(Y,U) \right\}. \qquad (5)$$

Thus if we can estimate $v(y,u)$ or its approximation $v_\beta(y,u)$ from sample paths, then we can estimate $\nabla \eta$ or $\nabla_\beta \eta$ using a sampling version of Eq. (3) or Eq. (5).

It turns out that by ergodicity of the Markov chain, we are able to compute $v_\beta$ from a sample trajectory, and compute $v$ with some bias. This was first noticed by (Jaakkola, Singh, and Jordan, 1994). Let us reason informally why it is so for the case of $v_\beta(y,u)$. Let $\pi(x,y,u)$ be the equilibrium distribution of the Markov chain $\{(X_t, Y_t, U_t)\}$, and $\pi(y,u)$, $\pi(x|y,u)$ be the corresponding marginal and conditional distributions, respectively. For any sample trajectory $\{(y_t, u_t)\}_{t \leq T}$, by ergodicity the number of the sub-trajectories that start with $(y,u)$, denoted by $T_{y,u}$, will be approximately $\pi(y,u)T$, as $T \to \infty$. Among these sub-trajectories the number of those that start from the state $x$ will be approximately $\pi(x|y,u)T_{y,u}$. Thus averaging over the discounted total costs of these $T_{y,u}$ sub-trajectories, we obtain in the limit $v_\beta(y,u)$, as $T \to \infty$.

Using ergodicity, one can have many ways of estimating $v(y,u)$ or $v_\beta(y,u)$ from sample paths. We will focus on the temporal difference methods in the following as they have well-established convergence and approximation error analysis.

## 2.3 COMPUTING $v_\beta(y, u)$ AND $v(y, u)$ BY TD ALGORITHMS

Let $\Phi$ be a matrix with rows $\phi(y, u)'$ – called the features of $(y, u)$, and with linearly independent columns – called basis functions, such that the column space includes the set of functions $\left\{\frac{1}{\mu_u(y,\theta)} \frac{\partial \mu_u(y,\theta)}{\partial \theta_1}, \ldots, \frac{1}{\mu_u(y,\theta)} \frac{\partial \mu_u(y,\theta)}{\partial \theta_k}\right\}$ which we call the minimum set of basis functions. We approximate the function $v(y, u)$ or $v_\beta(y, u)$ by $\phi(y, u)'r$, where $r$ is a vector of linear coefficients, to be computed by TD algorithms.[5]

The same as in MDP (Konda and Tsitsiklis, 1999), (Sutton et al., 1999), Eq. (3) or Eq. (5) shows that for gradient estimation, we only need to estimate the projection of the function $v(Y, U)$ or $v_\beta(Y, U)$, (viewed as a random variable), on a subspace that includes the minimum set of basis functions $\left\{\frac{1}{\mu_U(Y,\theta)} \frac{\partial \mu_U(Y,\theta)}{\partial \theta_1}, \ldots, \frac{1}{\mu_U(Y,\theta)} \frac{\partial \mu_U(Y,\theta)}{\partial \theta_k}\right\}$, (viewed as random variables), where the projection is with respect to the marginal equilibrium distribution $\pi(y, u)$.

We note that from this projection viewpoint, without resorting to $v$, the original gradient expression Eq. (1) is already sufficient for the claim that the state information is not necessary in biased gradient estimation, because the minimum set of basis functions are not functions of the state.

What we show in the following – combining established results on TD with an error decomposition – is that the two arguments are equivalent and complementary in the sense that (i) the value functions obtained by TD in the limit are (unbiased or biased) projections of $v$ or $v_\beta$, and (ii) any MDP with linear value function approximation can be viewed as a POMDP with the feature of a state as the observation. Our analysis completes the part of an earlier work (Singh, Jaakkola, and Jordan, 1994) on the same subject.

For clarity we define another set of features $\tilde{\phi}(x, y, u) = \phi(y, u)$. We run TD algorithms with features $\tilde{\phi}$ in the POMDP.

---

[5]TD algorithms include the original TD algorithms (e.g., (Sutton, 1988), (Bertsekas and Tsitsiklis, 1996), (Tsitsiklis and Van Roy, 1999)), the least squares TD algorithms (e.g., (Boyan, 1999), (Bertsekas, Bokar, and Nedić, 2003)), and many other variants. They differ in convergence rate and computation overhead, and they converge to the same limits.

**Estimation of $v_\beta$ by Discounted TD**

Consider the $\beta$-discounted TD($\lambda$) with $\lambda = 1$. Let $r_\beta^*$ be the limit of the linear coefficients that TD converges to, and $\hat{v}_\beta(y, u) = \phi(y, u)'r_\beta^*$ be the corresponding function.

**Proposition 1** *The function $\hat{v}_\beta(y, u)$ is a projection of $v_\beta(y, u)$ on the column space of $\Phi$, i.e.,*

$$E_0 \left\{ (v_\beta(Y, u) - \phi(Y, U)'r_\beta^*)^2 \right\}$$
$$= \min_{r \in \mathcal{R}^k} E_0 \left\{ (v_\beta(Y, u) - \phi(Y, U)'r)^2 \right\}.$$

Prop. 1 follows from results on discounted TD and the next simple lemma, which follows from the fact that $v_\beta(y, u)$ is the conditional mean of $Q_\beta$.

**Lemma 1** *For any vector $r \in \mathcal{R}^k$,*

$$E_0 \left\{ (Q_\beta(X, Y, U) - \phi(Y, U)'r)^2 \right\}$$
$$= E_0 \left\{ Q_\beta(X, Y, U)^2 - v_\beta(Y, U)^2 \right\}$$
$$+ E_0 \left\{ (v_\beta(Y, U) - \phi(Y, U)'r)^2 \right\}. \quad (6)$$

**Proof of Prop. 1:** Since $\lambda = 1$, by Proposition 6.5 in (Bertsekas and Tsitsiklis, 1996) (pp. 305), the function $\tilde{\phi}(x, y, u)'r_\beta^*$ is the projection of $Q_\beta(x, y, u)$ on the feature space with respect to the equilibrium distribution, i.e., $r_\beta^*$ minimizes

$$E_0 \left\{ \left( Q_\beta(X, Y, U) - \tilde{\phi}(X, Y, U)'r \right)^2 \right\}$$
$$= E_0 \left\{ (Q_\beta(X, Y, U) - \phi(Y, U)'r)^2 \right\}.$$

Hence $r_\beta^*$ minimizes $E_0 \left\{ (v_\beta(Y, U) - \phi(Y, U)'r)^2 \right\}$ by Lemma 1. □

The error analysis for the case of $\lambda < 1$, omitted here, is similar to and less complicated than the case of average cost TD($\lambda$) as shown next.

**Estimation of $v$ by Average Cost TD**

Consider the average cost TD($\lambda$) with $\lambda < 1$.[6] Let $r_\lambda^*$ be the limit of the linear coefficients that TD converges to, and $\hat{v}(y, u) = \phi(y, u)'r_\lambda^*$ be the corresponding function. The next proposition says that modulo a constant translation, $\hat{v}$ is an approximation to the projection of $v$ on the feature space, and converges to this projection when $\lambda \uparrow 1$. Its proof, given in the appendix,[7] is a straightforward combination of the results for average cost TD (Theorem 3 of (Tsitsiklis and Van Roy, 1999)) and a decomposition of error by Lemma 1.

---

[6]We assume that the column space of $\Phi$ does not contain the vector $[1 \ldots 1]'$, to satisfy a condition in average cost TD algorithms.

**Proposition 2** *There exists a constant scalar $\bar{c}$ such that*

$$E_0\left\{(v(Y,U) + \bar{c} - \phi(Y,U)'r_\lambda^*)^2\right\}$$
$$\leq \frac{\alpha_\lambda^2}{1-\alpha_\lambda^2} E_0\left\{Q(X,Y,U)^2 - v(Y,U)^2\right\}$$
$$+ \frac{1}{1-\alpha_\lambda^2} \inf_{r\in\mathcal{R}^k} \inf_{c\in\mathcal{R}} E_0\left\{(v(Y,U) + c - \phi(Y,U)'r)^2\right\},$$

*where $\alpha_\lambda \in [0,1)$ is a mixing factor, depending on the Markov chain, with $\lim_{\lambda\uparrow 1} \alpha_\lambda = 0$.*

By Prop. 2, the approximation error, measured in the squared norm, is bounded by two terms. The second term is a multiple of the best approximation error possible, and is zero when $v(y,u)$, modulo a constant translation, is in the feature space. The first term, vanishing as $\lambda \uparrow 1$, can be equivalently written as a multiple of the expectation of the variance of $Q(X,Y,U)$ conditioned on $(Y,U)$:

$$\frac{\alpha_\lambda^2}{1-\alpha_\lambda^2} E_0\left\{Var\left\{Q(X,Y,U) \mid Y,U\right\}\right\}$$

It does not depend on the features, and is a penalty for not observing states $X$.

## 3  GRADIENT ESTIMATION FOR FINITE-STATE CONTROLLERS

The graphical model of POMDP with a finite-state controller is shown in Fig. 2. The controller has an internal state, denoted by $Z_t$, taking a finite number of values. Given the observation $Y_t$, the controller applies the control $U_t$ with probability $p(U_t|Z_t,Y_t)$, and its internal state subsequently transits to $Z_{t+1}$ with probability $p(Z_{t+1}|Z_t,Y_t,U_t)$.[8] The process $\{(X_t,Y_t,Z_t,U_t)\}$ jointly forms a Markov chain, and so does the marginal process $\{(X_t,Y_t,Z_t)\}$.

Let $\{\gamma_\theta \mid \theta \in \Theta\}$ be a parametrized family of finite-state controllers with the same internal state space $Z$. For any policy $\gamma_\theta$, let

$$\mu_u(z,y,\theta) = p(U_t = u \mid Z_t = z, Y_t = y; \theta)$$

---
[7]See Footnote 3.

[8]One can define a finite-state controller different from the one we use here. For example, the internal state transits to $Z_{t+1}$ with probability $p(Z_{t+1}|Z_t,U_t,Y_{t+1})$, i.e., the transition depends on $Y_{t+1}$, instead of $Y_t$. The general idea outlined in Sec. 2 applies in the same way. The equations will be different from the ones in this section, however.

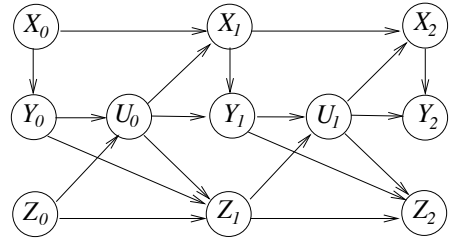

Figure 2: POMDP with a Finite-State Controller. The states $Z_t$ are the internal states of the controller.

be the probability of taking control $u$ at internal state $z$ and observation $y$, and

$$\zeta_{\bar{z}}(z,y,u,\theta) = p(Z_{t+1} = \bar{z} \mid Z_t = z, Y_t = y, U_t = u; \theta).$$

be the transition probability of the internal states. We require that $\mu_u(z,y,\theta)$ and $\zeta_{\bar{z}}(z,y,u,\theta)$ are differentiable for any give $u,z,y,\bar{z}$, and the transition probability $p(X_{t+1} = \bar{x}, Y_{t+1} = \bar{y}, Z_{t+1} = \bar{z} \mid X_t = x, Y_t = y, Z_t = z; \theta)$ is differentiable for any given $x,y,z,\bar{x},\bar{y},\bar{z}$. Similar to the case of reactive policies, we assume that for all $\theta \in \Theta$, the Markov chains $\{(X_t,Y_t,Z_t)\}$ can be defined on a common state space, and furthermore

**Assumption 3** *Under any policy $\gamma_\theta$, the Markov chain $\{(X_t,Y_t,Z_t)\}$ is irreducible and aperiodic.*

**Assumption 4** *There exists a constant L, such that for all $\theta \in \Theta$*

$$\max_{u,y}\left\|\frac{\nabla\mu_u(y,\theta)}{\mu_u(y,\theta)}\right\| \leq L, \quad \max_{u,y,z,\bar{z}}\left\|\frac{\nabla\zeta_{\bar{z}}(z,y,u,\theta)}{\zeta_{\bar{z}}(z,y,u,\theta)}\right\| \leq L,$$

*where $0/0$ is regarded as $0$.*

### 3.1  GRADIENT ESTIMATION

The gradient equals the sum of two terms:

$$\nabla\eta = E_0\left\{\frac{\nabla\mu_{U_0}(Z_0,Y_0,\theta)}{\mu_{U_0}(Z_0,Y_0,\theta)}Q(X_0,Y_0,Z_0,U_0)\right\}$$
$$+ E_0\left\{\frac{\nabla\zeta_{Z_1}(Z_0,Y_0,U_0,\theta)}{\zeta_{Z_1}(Z_0,Y_0,U_0,\theta)}h(X_1,Y_1,Z_1)\right\}, \quad (7)$$

where the Q-function $Q(x,y,u)$ is defined by

$$Q(x,y,z,u) = g(x,y,u)+$$
$$E\{h(X_1,Y_1,Z_1) \mid (X_0,Y_0,Z_0,U_0) = (x,y,z,u)\},$$

and $h(\cdot)$ is the bias function of policy $\gamma_\theta$.

To estimate the first term of the r.h.s. of Eq. (7), we can write the term as

$$E_0\left\{\frac{\nabla\mu_{U_0}(Z_0,Y_0,\theta)}{\mu_{U_0}(Z_0,Y_0,\theta)}v_1(Y_0,Z_0,U_0)\right\}$$

where
$$v_1(Y_0, Z_0, U_0) = E_0\{Q(X_0, Y_0, Z_0, U_0) \mid Y_0, Z_0, U_0\}$$

For estimating $v_1$, consider the Markov chain $\{(X_t, Y_t, Z_t)\}$, and apply $\beta$-discounted or average cost TD algorithms with the features $\tilde\phi(x, y, z, u) = \phi(y, z, u)$ not depending on $x$.

To estimate the second term of the r.h.s. of Eq. (7), we first note the relationship between the bias function $h(x, y, z)$ of the Markov chain $\{(X_t, Y_t, Z_t)\}$ and the bias function of the Markov chain $\{(X_t, Y_t, Z_t, U_t, Z_{t+1})\}$, denoted by $\tilde h(x, y, z, u, \bar z)$:

$$h(x, y, z) = \\ E\left\{\tilde h(x, y, z, U_0, Z_1) \mid (X_0, Y_0, Z_0) = (x, y, z)\right\},$$

which can be verified from the optimality equations. It follows that

$$\begin{aligned}&E\{h(X_1, Y_1, Z_1) \mid X_0, Y_0, Z_0, U_0, Z_1\}\\ &= E\left\{\tilde h(X_1, Y_1, Z_1, U_1, Z_2) \mid X_0, Y_0, Z_0, U_0, Z_1\right\}\\ &= \tilde h(X_0, Y_0, Z_0, U_0, Z_1) + \eta - g(X_0, Y_0, U_0)\end{aligned}$$

and since $\eta - g(X_0, Y_0, U_0)$ does not depend on $Z_1$, it can be dropped in gradient estimation. Hence the second term in the gradient expression equals

$$\begin{aligned}&E_0\left\{\frac{\nabla\zeta_{Z_1}(Z_0, Y_0, U_0, \theta)}{\zeta_{Z_1}(Z_0, Y_0, U_0, \theta)} \tilde h(X_0, Y_0, Z_0, U_0, Z_1)\right\}\\ &= E_0\left\{\frac{\nabla\zeta_{Z_1}(Z_0, Y_0, U_0, \theta)}{\zeta_{Z_1}(Z_0, Y_0, U_0, \theta)} v_2(Y_0, Z_0, U_0, Z_1)\right\}\end{aligned}$$

where

$$v_2(Y_0, Z_0, U_0, Z_1) = \\ E_0\left\{\tilde h(X_0, Y_0, Z_0, U_0, Z_1) \mid Y_0, Z_0, U_0, Z_1\right\}.$$

For estimating $v_2$, consider the Markov chain $\{(X_t, Y_t, Z_t, U_t, Z_{t+1})\}$, and apply the TD algorithms with the features $\tilde\phi(x, y, z, u, \bar z) = \phi(y, z, u, \bar z)$ not depending on $x$. The line of error analysis in Sec. 2.3 applies in the same way here.

## 4 GRADIENT ESTIMATION FOR POSMDP

POSMDPs stands for partially observable semi-Markov decision processes. Analogous to POMDP, we define POSMDPs as semi-Markov decision processes (SMDPs) with hidden states and observations generated by states. The model of SMDP is the same as MDP except that the time interval $\tau_{n+1} - \tau_n$, called the *sojourn time*, between transition from state $X_n$ at time $\tau_n$ to state $X_{n+1}$ at time $\tau_{n+1}$, is random, and depends on $X_n$, $X_{n+1}$ and the applied control $U_n$. The random variables $\{\tau_n\}$, called *decision epochs*, are the only time when controls can be applied. For details of SMDP, see (Puterman, 1994).

We consider the problem of POSMDP with a *subset* of finite state controllers that take the observations, but *not* the sojourn times, as inputs. This is to preserve the SMDP structure of the joint process $\{(X_n, Y_n, Z_n, U_n)\}$ and the marginal process $\{(X_n, Y_n, Z_n)\}$. (Singh, Tadic, and Doucet, 2002) gave a GPOMDP type gradient estimation algorithm for this problem. We would like to point out that the function approximation approach applies as well. The details are as follows.

The average cost is defined as the limit of the expected cost up to time $T$ divided by $T$, and under the irreducibility condition of the Markov chain $\{(X_n, Y_n, Z_n)\}$, by ergodicity the average cost equals to

$$\eta = \frac{E_0\{g(X_0, Y_0, U_0)\}}{E_0\{\tau_1\}},$$

where $g(x, y, u)$ is the mean of the random per-stage cost $c(x, y, \tau, u)$ that depends on the sojourn time $\tau$. In the case of reactive policies, one can show that the gradient equals to

$$\nabla\eta = \frac{E_0\left\{\frac{\nabla\mu_{U_0}(Y_0, \theta)}{\mu_{U_0}(Y_0, \theta)} h(X, Y, U)\right\}}{E_0\{\tau_1\}}$$

where $h$ satisfies the equation

$$\begin{aligned}h(x, y, u) =\ &g(x, y, u) - \bar\tau(x, y, u)\eta\\ &+ E\{h(X_1, Y_1, U_1) \mid (X_0, Y_0, U_0) = (x, y, u)\},\end{aligned}$$

and $\bar\tau(x, y, u)$ is the expected sojourn time given $(X_0, Y_0, U_0) = (x, y, u)$.

Now notice that $h$ is the bias function of the Markov chain $\{(X_n, Y_n, U_n)\}$ with $g(x, y, u) - \bar\tau(x, y, u)\eta$ as the expected per-stage cost, or equivalently with $c(X, Y, \tau, U) - \tau(X, Y, U)\eta$ as the random per-stage cost, where $\tau$ is the random sojourn time. Let $\hat\eta_n$ be the online estimate of $\eta$. We can thus estimate the projection of $h$ (equivalently the conditional mean of $h$) by running TD algorithms (discounted or average cost version) in this MDP with per-stage cost $g_n - (\tau_{n+1} - \tau_n)\hat\eta_n$, and with features not depending on state $x$ and sojourn time $\tau$.

The general case of finite-state controllers is similar: the gradient equals the sum of two parts, each of which can be estimated using function approximation by considering the appropriate Markov chain – the same as in POMDP – with per-stage cost $g_n - (\tau_{n+1} - \tau_n)\hat\eta_n$.

## 5 EXPERIMENTS

We test GPOMDP and our method on a medium size ALOHA problem – a communication problem — with 30 states, 3 observations and 9 actions.[9] We take its model from A. R. Cassandra's POMDP data repertoire (on the web), and define per-stage costs to be the negative rewards. The true gradients and average costs in comparison are computed using the model. The family of policies we used has 3 internal states, 72 action parameters governing the randomized control probabilities $\mu_u(z, y, \theta)$, and 1 internal-transition parameter governing the transition probabilities of the internal states $\zeta_{\bar{z}}(z, y, u, \theta)$.[10] The parameters are bounded so that all the probabilities are in the interval $[0.001, 0.999]$. For experiments reported below, $\beta = 0.9, \lambda = 0.9$.

We demonstrate below the behavior of gradient estimators in two typical situations: when the magnitude of the true gradient is large, and when it is small. Correspondingly they can happen when the policy parameter is far away from a local minima, and when it is close to a local minima (or local maxima).

First we describe how the local minima was found, which also shows that the approach of finite-state controller with policy gradient is quite effective for this problem. The initial policy has equal action probabilities for all internal-state and observation pairs, and has 0.2 as the internal-transition parameter. At each iteration, the gradient is estimated from a simulated sample trajectory of length 20000 (a moderate number for the size of this problem), emphwithout using any estimates from previous iterations. We then, denoting the estimate by $\hat{\nabla}\eta$, project $-\hat{\nabla}\eta$ to the feasible direction set, and update the policy parameter by a small constant step along the projected direction. We used GPOMDP in this procedure, (mainly because it needs less computation). The initial policy has average cost $-0.234$. The cost monotonically decreases, and within 4000 iterations the policy gets into the neighborhood of a local minima, oscillating around afterwards, with average costs in the interval $[-0.366, -0.361]$ for the last 300 iterations. As a comparison, the optimal (liminf) average cost of this POMDP is bounded below by $-0.460$, which is computed using an approximation scheme from (Yu and Bertsekas, 2004).

Table 1: Comparison of Gradient Estimators. The number $\frac{\hat{\nabla}\eta' \nabla\eta}{\|\hat{\nabla}\eta\|_2 \|\nabla\eta\|_2}$ when $\theta$ is far away from a local minima.

| B-TD | OL-TD | GPOMDP |
|---|---|---|
| $0.9678 \pm 0.0089$ | $0.875 \pm 0.006$ | $0.9680 \pm 0.0088$ |

Table 1 lists the number $\frac{\hat{\nabla}\eta' \nabla\eta}{\|\hat{\nabla}\eta\|_2 \|\nabla\eta\|_2}$ for several gradient estimators, when the policy is far from a local minima. The values listed are the means and standard deviations calculated from 5 sample trajectories simulated under the same policy. In the first column, the gradient estimator (B-TD) uses the batch estimate of the value function, that is, it uses the function estimated by TD at the end of a trajectory. In the second column, the gradient estimator (OL-TD) uses the on-line estimates of the value function computed by TD. The TD algorithms we used are $\beta$-discounted LSPE($\lambda$) (Bertsekas, Bokar, and Nedić, 2003) and average cost LSPE($\lambda$). The difference between the discounted and average cost TD turns out negligible in this experiment. In the third column, we use GPOMDP.[11] The estimates from B-TD and GPOMDP align well with the true gradient, while OL-TD is not as good, due to the poor estimates of TD in the early period of a trajectory.

Fig. 3 shows the number $\frac{\hat{\nabla}\eta' \nabla\eta}{\|\hat{\nabla}\eta\|_2 \|\nabla\eta\|_2}$ for several gradient estimators on 20 sample trajectories simulated under the same policy, when that policy is near a local minima.[12] The horizontal axis indexes the trajectories. The blue solid line and the green dash-dot line correspond, respectively, to the gradient estimator that uses the batch estimate (B-TD) and the on-line estimate (OL-TD) of the value function, computed by $\beta$-discounted LSPE($\lambda$). The red dash line corresponds to GPOMDP. While the estimator B-TD consistently aligns well with the true gradient, GPOMDP often points to the opposite direction.

Our experiments demonstrate that when close to a local minima (or local maxima), where the magnitude of the gradient is small, in order to align with the gradient, the estimator needs to have much smaller bias and variance. In GPOMDP we only have one parameter $\beta$ to balance the bias-variance. Hence it can be advantageous for the function approximation approach to provide more options – namely the feature space, $\lambda$ and $\beta$ – in controlling bias and variance in gradient estimation.

---

[9]In this problem, a state generates the same observation under all actions, and for each observation, the number of states that can generate it is 10.

[10]The internal-transitions are made such that the internal-state functions as a memory of the past, and the parameter is the probability of remembering the previous internal-state, with 1 minus the parameter being the probability of refreshing the internal state by the recent observation.

[11]For both GPOMDP and the discounted TD algorithm, we subtracted the per-stage cost by the on-line estimate of the average cost.

[12]More precisely, the number we compute here is the inner-product of the *projections* of $-\nabla\eta$ and $-\hat{\nabla}\eta$ (on the set of feasible directions) normalized by their norms.

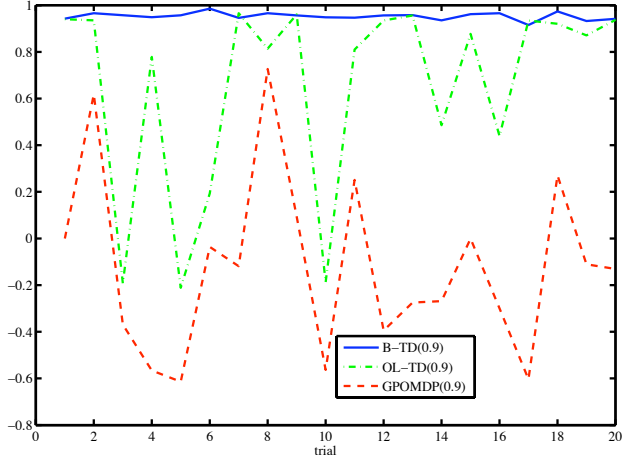

Figure 3: Comparison of Gradient Estimators. The number $\frac{\hat{\nabla}\eta'\nabla\eta}{\|\hat{\nabla}\eta\|_2\|\nabla\eta\|_2}$ when $\theta$ is near a local minima. Linear interpolations between trials are plotted for reading convenience.

## 6 Discussion

We have shown that Actor-Critic methods are alternatives to GPOMDP in learning finite-state controllers for POMDP and POSMDP. Actor-Critic methods provide more options in bias-variance control than GPOMDP. It is unclear, however, both theoretically or practically, which method is most efficient: actor-only, actor-crictic, or their combined variants as suggested in (Konda, 2002). We also note that using a value function in gradient estimation can be viewed as a variance reduction technique based on Rao-Blackwellization. The control variate idea (Greensmith, Bartlett, and Baxter, 2004) is a different type of variance reduction technique, and applies to both actor-only and actor-critic algorithms.

**Acknowledgments**

This work is supported by NSF Grant ECS-0218328. The author would like to thank Prof. D. P. Bertsekas and the anonymous reviewers for helpful feedback.